%% file: IEEE-conference-template-062824.tex
\documentclass[conference]{IEEEtran}
\IEEEoverridecommandlockouts
\usepackage{cite}
\usepackage{amsmath,amssymb,amsfonts}
\usepackage{algorithmic}
\usepackage{graphicx}
\usepackage{textcomp}
\usepackage{caption}
\usepackage{color, xcolor}
\usepackage{booktabs}
\usepackage{multirow}
\usepackage[capitalise]{cleveref}
\def\BibTeX{{\rm B\kern-.05em{\sc i\kern-.025em b}\kern-.08em
    T\kern-.1667em\lower.7ex\hbox{E}\kern-.125emX}}
    
\begin{document}

\title{Bits for Privacy: Evaluating Post-Training Quantization via Membership Inference}

\author{
\IEEEauthorblockN{1\textsuperscript{st} Chenxiang Zhang}
\IEEEauthorblockA{
\textit{Department of Computer Science} \\
\textit{University of Luxembourg}\\
Esch-sur-Alzette, Luxembourg \\
chenxiang.zhang@uni.lu}
\and
\IEEEauthorblockN{2\textsuperscript{nd} Tongxi Qu}
\IEEEauthorblockA{\textit{Department of Computer Science} \\
\textit{Nanjing University}\\
Nanjing, China \\
tongxiqu@smail.nju.edu.cn}
\and
\IEEEauthorblockN{3\textsuperscript{rd} Zhong Li}
\IEEEauthorblockA{\textit{Department of Computer Science} \\
\textit{Nanjing University}\\
Nanjing, China \\
lizhong@nju.edu.cn}
\and
\IEEEauthorblockN{4\textsuperscript{th} Tian Zhang}
\IEEEauthorblockA{\textit{Department of Computer Science} \\
\textit{Nanjing University}\\
Nanjing, China \\
ztluck@nju.edu.cn}
\and
\IEEEauthorblockN{5\textsuperscript{th} Jun Pang}
\IEEEauthorblockA{\textit{Department of Computer Science} \\
\textit{University of Luxembourg}\\
Esch-sur-Alzette, Luxembourg \\
jun.pang@uni.lu}
\and
\IEEEauthorblockN{6\textsuperscript{th} Sjouke Mauw}
\IEEEauthorblockA{\textit{Department of Computer Science} \\
\textit{University of Luxembourg}\\
Esch-sur-Alzette, Luxembourg \\
sjouke.mauw@uni.lu}

}

\maketitle

\begin{abstract}
Deep neural networks are widely deployed with quantization techniques to reduce memory and computational costs by lowering the numerical precision of their parameters. 
While quantization alters model parameters and their outputs, existing privacy analyses primarily focus on full-precision models, leaving a gap in understanding how bit-width reduction can affect privacy leakage.
We present the first systematic study of the privacy–utility relationship in post-training quantization (PTQ), a versatile family of methods that can be applied to pretrained models without further training.
Using membership inference attacks as our evaluation framework, we analyze three popular PTQ algorithms—AdaRound, BRECQ, and OBC—across multiple precision levels (4-bit, 2-bit, and 1.58-bit) on CIFAR-10, CIFAR-100, and TinyImageNet datasets.
Our findings consistently show that low-precision PTQs can reduce privacy leakage. In particular, lower-precision models demonstrate up to an order of magnitude reduction in membership inference vulnerability compared to their full-precision counterparts, albeit at the cost of decreased utility.
Additional ablation studies on the 1.58-bit quantization level show that quantizing only the last layer at higher precision enables fine-grained control over the privacy-utility trade-off. 
These results offer actionable insights for practitioners to balance efficiency, utility, and privacy protection in real-world deployments.
\end{abstract}

\smallskip
\begin{IEEEkeywords}
neural network, membership inference, quantization, post-training quantization
\end{IEEEkeywords}

%%%%%%%%%%%%%%%%%%%%%%%%%%%%%%%%%%%%%%%%%%%%%%%%%%%%%%%%%%%%%%%%%%%%%%%%
\section{Introduction}
\label{sec:intro}

\begin{figure}[!t]
    \centering
    \includegraphics[width=0.75\linewidth]{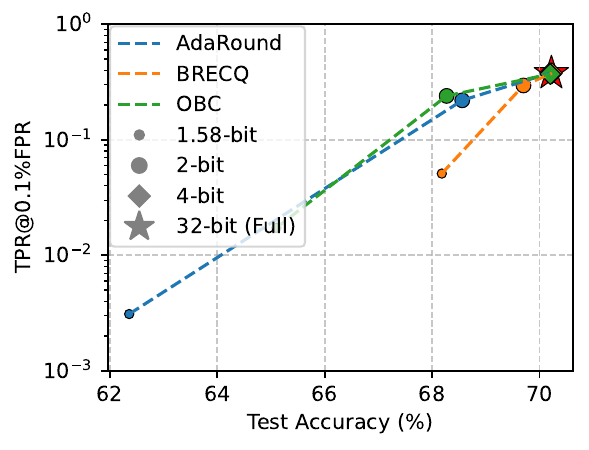}
    \caption{Privacy leakage of quantized models trained on CIFAR-100 under LiRA online attack. Lower numerical precision has a worse utility but better privacy protection evaluated with TPR@0.1\%FPR. Results are averaged across 10 random seeds. }
    \label{fig:teaser}
\end{figure}

\smallskip\noindent
As neural networks scale to millions and billions of parameters, their deployment faces significant computational and memory constraints.
Quantization addresses this challenge by reducing the storage and computational costs of neural networks through lowering parameter precision (e.g., converting 32-bit floating-point numbers to 8-bit integers). 
Given these practical reasons, quantization approaches have been widely applied and studied~\cite{hubara2018quantized,le2023efficient} in recent years to enable more efficient model inference, often with minimal loss in accuracy~\cite{gholami2022survey}. 
Quantization methods are classified into two categories: Quantization-Aware Training (QAT)~\cite{nagel2021white} and Post-Training Quantization (PTQ). The methods in the first category directly train a quantized model from scratch, while the second applies quantization to a pre-trained model at full-precision using a small set of data for calibration. 
% PTQ methods are easier to use and therefore they are more widely adopted in practice.

While quantization methods have been shown to successfully improve computational efficiency, they may introduce undesired side effects. By altering parameter distributions, quantization can disproportionately degrade performance on difficult samples, leading to fairness concerns~\cite{hooker2019compressed,hooker2020characterising}. 
These changes to parameters can also have privacy implications. 
Recent work suggests that parameter modifications from compression can influence the extent of sensitive information leakage under privacy attacks~\cite{towards2022model,jin2023membership}. 
Although these studies indicate that simple uniform quantization can reduce privacy leakage, the privacy of more sophisticated methods such as PTQ remain unexplored. This gap is particularly interesting given PTQ's adoption in practice. 

In this paper, we systematically assess the potential privacy risks introduced by PTQ using Membership Inference Attacks (MIAs) as our evaluation tool for privacy threats. MIAs aim to determine whether a given data sample was used in a model’s training process~\cite{shokri2017membership,yeom2018privacy,HP23}. 
% These attacks exploit the differences in model behavior between training and non-training data, typically by analyzing confidence scores, loss values, or internal representations~\cite{nasr2019comprehensive,song2021systematic,carlini2022membership}. 
A model highly vulnerable to MIAs may inadvertently leak sensitive information about its training data, indicating weaker privacy protection. Therefore, MIAs have been widely adopted as a metric to assess the privacy risks of a model~\cite{dnn2023quantization}. 

\smallskip\noindent
{\bf Contributions.}
We investigate the privacy leakage of PTQ methods~\cite{gholami2022survey}. In contrast to prior works, we employ state-of-the-art Likelihood Ratio Attack (LiRA)~\cite{carlini2022membership} membership attack and evaluate the attack success using the true positive rate (TPR) under a low false positive rate (FPR). 
LiRA computes a data instance-based score, overcoming the limitations of the global threshold attack~\cite{yeom2018privacy} which assumes that all the data points are similar 
We assess the privacy leakage associated with three widely used post-training quantization methods: AdaRound~\cite{nagel2020up}, BRECQ~\cite{li2021brecq}, and OBC~\cite{frantar2022optimal}. For each method, we evaluate and compare their utility performance and the privacy leakage at different quantization levels: 32-bit, 4-bit, 2-bit, and 1.58-bit~\cite{ma2024era}. 
~\Cref{fig:teaser} summarizes our main findings, where we identify a clear trade-off between utility and privacy for PTQ methods applied to deep neural networks. For 4-bit quantization, we achieve comparable accuracy as the full-precision model on the CIFAR-10, CIFAR-100~\cite{krizhevsky2009learning}, and TinyImageNet~\cite{le2015tiny} datasets. At a lower precision, 2-bit, the models have lower accuracy but also lower privacy leakage. Lastly, with an even more extreme 1.58-bit quantization, the trade-off between accuracy and privacy is far more noticeable. Specifically, the privacy leakage of the quantized models is an order of magnitude lower than the full-precision model.  

Further ablation results in ~\Cref{sec:ablation} demonstrate that our results hold across different architecture families and scales. We also find that ``decoupled quantization'', where we quantize the last layer with a higher-bit than others, can be used to control the utility-privacy trade-off at 1.58-bit quantization. Specifically, for TinyImageNet, decoupling can almost recover the full model accuracy while maintaining a low TPR@0.1\%FPR. 
% We release our code publicly at: \url{https://anonymous.4open.science/r/Quant-MIA-20E0}.

%%%%%%%%%%%%%%%%%%%%%%%%%%%%%%%%%%%%%%%%%%%%%%%%%%%%%%%%%%%%%%%%%%%%%%%%

\section{Preliminaries}
\label{sec:pre}

\noindent
A {\bf Deep Neural Network} (DNN) is a function \( f_\theta\colon \mathbb{R}^d \rightarrow \mathbb{R}^k \), where \( \theta \) denotes the model parameters (weights and biases), \( d \) is the input dimension, and \( k \) is the number of output classes. A typical DNN is composed of multiple layers of affine transformations followed by non-linear activations, defined as:
\begin{equation}
h^{(\ell)} = \sigma(W^{(\ell)} h^{(\ell-1)} + b^{(\ell)}), \quad \ell = 1, 2, \dots, L    
\end{equation}
where \( h^{(0)} = x \in \mathbb{R}^d \), \( W^{(\ell)} \in \mathbb{R}^{n_\ell \times n_{\ell-1}} \), \( b^{(\ell)} \in \mathbb{R}^{n_\ell} \), and \( \sigma(\cdot) \) denotes a non-linear function.

\smallskip\noindent
A {\bf Uniform Affine Quantization} (UAQ) maps continuous real numbers (e.g., represented by 32-bit floating point values) to discrete integers within a fixed range. To enable efficient storage and computation on computers, asymmetric quantization typically selects \( Q = \{0, 1, \dots, 2^b - 1\} \), where \( b \) is the number of bits used for quantization.  Specifically, for \( b \)-bit UAQ, given a chosen scale factor \( s \) and zero-point \( z \) based on the parameter range, after the quantization process, the parameter \(\hat{W}\) is described by the following equation:  

\begin{equation}
\label{eq:quantization}
\hat{W} = s\times\left(\text{clamp}\left(\left\lfloor \frac{W}{s} \right\rceil + z, 0, 2^b - 1\right) - z\right) 
\end{equation}
where the clamp function is defined as:  
\begin{equation}
\quad \text{clamp}(a, b, c) = \max(b, \min(a, c))  
\end{equation}

Since UAQ uniformly maps values to integer grid points, and the selection of the scale factor \( s \) is easy to implement, UAQ is computationally efficient and hardware-friendly. Many widely used PTQ methods are built upon adjustments to UAQ~\cite{nagel2020up, li2021brecq, frantar2022optimal}. 

\smallskip\noindent
{\bf Membership Inference Attacks} (MIAs) represent the simplest privacy threat to machine learning models by determining whether a particular data point \( x \) was used during training. Formally, given a model \( f_\theta \) and query access to its predictions, an adversary attempts to infer the membership status \( m \in \{0,1\} \), where \( m = 1 \) indicates that \( x \in \mathcal{D}_{\text{train}} \), and \( m = 0 \) otherwise~\cite{shokri2017membership}.

DNNs have strong performance across various domains and applications, including computer vision and language understanding. However, they are prone to memorization of training data~\cite{zhang2017understanding}, making them also vulnerable to privacy attacks such as MIAs~\cite{shokri2017membership}.
% MIAs exploit the observation that deep neural networks often exhibit distinguishable behavior on training samples compared to unseen inputs. 
% Based on these signals, an attacker can leverage various statistics, such as prediction confidence, loss, or entropy, as membership indicators. In some settings, more advanced features are extracted, including gradient norms or internal activations~\cite{nasr2019comprehensive}. A common approach, introduced by~\cite{shokri2017membership}, involves training multiple shadow models that simulate the target model's behavior, then using their outputs to train a binary classifier that distinguishes members from non-members.

%%%%%%%%%%%%%%%%%%%%%%%%%%%%%%%%%%%%%%%%%%%%%%%%%%%%%%%%%%%%%%%%%%%%%%
\section{Post-Training Quantization Effect on Privacy}
\label{sec:quant}

\smallskip\noindent
We motivate our study and introduce the research questions. Then, we describe the different PTQ methods and the MIA used in our methodology.

\subsection{Motivation and research questions}

\smallskip\noindent
Prior studies have extensively investigated PTQ in the context of model efficiency and accuracy, but its implications for privacy remain underexplored. We proceed to highlight three key questions that drive our study.

\smallskip\noindent
{\bf Can PTQs improve model privacy?}
Post-training quantization is widely adopted in practical deployments thanks to its convenience of being applied directly to a pretrained model. If PTQ can also protect against MIAs, it can enhance both efficiency and privacy. To evaluate this possibility, we systematically assess three different state-of-the-art PTQ methods: AdaRound, BRECQ, and OBC. We evaluate the privacy leakage of quantized models using the TPR under low FPR obtained from the LiRA attack.

\smallskip\noindent
{\bf What is the cost of privacy protection?}
Privacy-enhancing techniques often come with significant utility degradation. For instance, while DP-SGD offers formal guarantees, the strict utility degradation renders it impractical in many practical scenarios~\cite{abadi2016deep}. We conduct a detailed analysis of the costs and results of mitigating MIA using PTQ by measuring both model accuracy and the TPR of the attack. This provides practical insight into the privacy-utility trade-off.

\smallskip\noindent
{\bf Can we control the privacy utility trade-off?}
PTQ methods such as BRECQ~\cite{li2021brecq} improve accuracy by selectively preserving higher precision for the first and last layers. Inspired by this, as demonstrated in~\Cref{sec:ablation}, we show that decoupling the last-layer to a higher precision (e.g. 8-bit) can lead to a significant gain in model accuracy while still maintaining comparable privacy levels.

\subsection{Quantization and MIA methods}

\smallskip\noindent
{\bf AdaRound}~\cite{nagel2020up} addresses the limitations of standard rounding in post-training quantization by learning adaptive rounding decisions while minimizing the impact on the model's output. Unlike naive round-to-nearest, which is input-independent and may lead to suboptimal quantization, AdaRound introduces a self-study, layer-wise optimization process. It parameterizes the rounding operation using a rectified sigmoid, allowing the rounding behavior to be relaxed during optimization and later discretized. By minimizing a local loss that approximates the impact of quantization on the layer's output, AdaRound learns whether each weight should be rounded up or down. This adaptive scheme significantly reduces reconstruction error and can be applied sequentially across layers to improve the overall accuracy of quantized models.

\smallskip\noindent
{\bf BRECQ}~\cite{li2021brecq} aims to improve the accuracy by reducing the second-order error. Noting that the layers in a neural network are not completely independent, BRECQ analyzes the cross-layer dependency and argues that the traditional approach of quantizing each layer separately is suboptimal. Instead, they reconstruct layers with strong dependencies into blocks and perform quantization on each block. They justify their approach both theoretically and experimentally. In the experimental part, to achieve higher performance, BRECQ adopts AdaRound as the base quantization method for the blocks. The block mechanism allows multiple layers to compensate for the quantization loss together, achieving higher accuracy compared to directly applying AdaRound at extreme low-bit quantization.

\smallskip\noindent
{\bf OBC}~\cite{frantar2022optimal} integrates both quantization and pruning based on the classic model pruning framework OBS. Frantar et al., 2022 \cite{frantar2022optimal} propose that the quantization process can be viewed as a special form of pruning. Specifically, instead of setting the pruned weight values to 0, the weight values to be pruned are assigned according to the quantization equation. After ensuring that a similar ``pruning'' operation is applied to all weights, OBC achieves the quantization procedure. The modified OBS can then be used for quantization. The quantization framework of OBC\footnote{We use ``OBC'' to refer specifically to the quantization method.} first applies OBS to greedily select the weights with the least impact on the current quantization error. After handling outliers, the selected weights are quantized by UAQ, and the Hessian matrix of the weights is updated. The selected weights are marked to avoid repeated computations. In this manner, OBC quantizes each weight sequentially.

% These three PTQs share the following advantages, making them widely used and suitable for our experiments: they maintain high accuracy in low-bit quantization, and they require minimal training data with low quantization overhead.

\smallskip\noindent
{\bf Likelihood Ratio Attack (LiRA)} is a membership inference attack based on likelihood ratio testing. The \textit{Online} attack trains multiple shadow models, one half including the queried data point and the other half excluding it, to estimate the distribution of model losses as a Gaussian distribution. This method achieves strong performance but is computationally expensive, as each query requires training new shadow models. To address this, an \textit{Offline} attack is also introduced, which uses only shadow models that exclude the query, trading the attack success for efficiency. Both variants support a \textit{fixed variance} mode, where the mean and standard deviation are averaged across shadow models to reduce estimation noise. This approximation improves performance when the number of shadow models is small.

\smallskip\noindent
{\bf 1.58-bit Post-Training Quantization.}
Prior quantization works have demonstrated that large-scale language models (LLM) can be effectively quantized to extremely low-bit-widths. In particular, Ma et al., 2024~\cite{ma2024era} propose BitNet b1.58, a ternary quantization-aware training method that constrains the weights to $\{-1, 0, 1\}$, corresponding to an average of $\log_2 3 \approx 1.58$ bits per parameter. This approach reduces computational and memory costs while maintaining comparable performance. The 1.58-bit quantization allows for efficient inference by replacing multiplications with additions, and the inclusion of zero naturally introduces sparsity, offering additional computation speed-up. 

Inspired by Ma et al., 2024, we extend the idea of 1.58-bit quantization to PTQ. For a weight parameter $W$, given a suitable scaling factor $s$ and zero point $z$, a 1.58-bit asymmetric uniform quantization can be formulated as follows:
\begin{equation}
\hat{W} = s \times \left( \text{clamp}\left( \left\lfloor \frac{W}{s} \right\rceil + z, 0, 2 \right) - z \right),
\label{eq:log3quant}
\end{equation}
where the clamp function ensures the quantized values remain within the range $\{0, 1, 2\}$. In matrix multiplication, a value of 0 is equivalent to skipping the operation, 1 corresponds to an addition, and 2 can be interpreted as a left-shift operation, which is similar to the results of Ma et al., 2024~\cite{ma2024era}. Based on this, it is possible to design a specialized computation mechanism to efficiently simulate the multiplication between a floating-point matrix and a matrix with only 0, 1, and 2. This approach can speed up the multiplication between a floating-point matrix and an integer matrix. Therefore, compared to 2-bit quantization, 1.58-bit quantization can indeed improve computational efficiency.

A key advantage of~\eqref{eq:log3quant} is that it can be integrated with the aforementioned PTQ methods. Noting that AdaRound replaces the $\lfloor\cdot\rceil$ operator with adaptive rounding, BRECQ only modifies the structure of the model, and OBC can also determine the quantization order according to~\eqref{eq:log3quant}, it is therefore feasible to apply the proposed 1.58-bit quantization in conjunction with these techniques. Although only three values are available for quantization, there remains sufficient flexibility to adjust the model's weights.

%%%%%%%%%%%%%%%%%%%%%%%%%%%%%%%%%%%%%%%%%%%%%%%%%%%%%%%%%%%%%%%%%%%%%%%%
%%%%%%%%%%%%%%%%%%%%%%%%%%%%%%%%%%%%%%%%%%%%%%%%%%%%%%%%%%%%%%%%%%%%%%%%%

\section{Experimental Results}
\label{sec:results}

\noindent
In this section, we first introduce the experiment settings, including the datasets, model, quantization methods, and privacy attack in \Cref{sec:setup}. 
Then, in~\Cref{sec:attack effectiveness}, we present the main results, describing the trade-off between quantization-utility-privacy. Furthermore, an ablation study on 1.58-bit quantization is carried out in~\Cref{sec:ablation}, where we set the last layer to a higher precision (e.g. 8-bit) to control this trade-off and study the accuracy-TPR distribution of quantized models to provide a detailed and practical guideline. 

\subsection{Experimental setup}
\label{sec:setup}

\noindent
{\bf Dataset \& Model.}
We include the commonly used vision benchmarks to assess privacy risks that have been widely adopted in previous MIA research~\cite{shokri2017membership,tang2022mitigating}.

\begin{itemize}
    \item \text{CIFAR-10}~\cite{krizhevsky2009learning} consists of 60,000 color images in 10 classes, with 6,000 images per class. Each image has a resolution of $32 \times 32$ pixels. The dataset is divided into 50,000 training images and 10,000 testing images.
    \item \text{CIFAR-100}~\cite{krizhevsky2009learning} shares the same resolution and number of images as CIFAR-10. Note, however, that CIFAR-100 is made of different images, and it is a more challenging task of 100 classes, with 600 images per class.
    \item \text{TinyImageNet}~\cite{le2015tiny} is a subset of the ILSVRC 2012 ImageNet dataset~\cite{russakovsky2015imagenet}, containing 200 object classes. Each class has 500 training, 50 validation, and 50 test images. Images are resized to $64 \times 64$ pixels.
\end{itemize}

Unless stated otherwise, we employ ResNet18~\cite{he2016deep} for the majority of our experiments. For all training data, we apply standard data augmentation techniques, such as random horizontal flipping, random cropping, and normalization. For the test data, only normalization is applied.  
Following the same experimental setup of Carlini et al., 2022~\cite{carlini2022membership}, we train the models for 100 epochs using 50\% of the training dataset to obtain the full-precision model. We use stochastic gradient descent (SGD) with a momentum of 0.9, a weight decay of 5e-4, a learning rate of 0.1 with a cosine annealing scheduler.

\smallskip\noindent
{\bf Quantization.}
When evaluating the privacy of quantized models, we must avoid data leakage during the quantization process. To this end, we have to ensure that the ``calibration'' data is a subset of the dataset used for training the full-precision model. 
As the name suggests, the calibration dataset is used to calibrate and evaluate the performance of the quantized model. 
We use 1024 samples selected randomly but fixed across different quantization procedures to ensure fair comparison. 
We perform per-channel asymmetric weight quantization only to the weights (the activation function remains full-precision). We evaluate three PTQ methods: AdaRound, BRECQ, and OBC. For OBC, we apply batchnorm tuning after quantization as in Frantar et al., 2022~\cite{frantar2022optimal}. 
Additional quantization parameters are kept consistent with the original implementations of the respective methods.
We quantize all the PTQ models to varying precision levels, including 1.58-bit, 2-bit, and 4-bit.

\smallskip\noindent
{\bf Membership Inference Attack.}
We evaluate the privacy leakage using three different attacks: LiRA online and LiRA offline\cite{carlini2022membership} with the fixed variance. We train 64 shadow models to evaluate the privacy leakage of models quantized using different PTQs at different precision levels, including full-precision, 4-bit, 2-bit, and 1.58-bit. 
The model training recipe follows the same setting as in Carlini et al., 2022~\cite{carlini2022membership}. That is, we ensure that each training data is included in exactly half of the shadow training datasets. 
We compare utility using the accuracy metric and privacy using the log-AUROC and TPR@0.1\%FPR, which captures a meaningful privacy threat instead of accuracy~\cite{carlini2022membership}. 

\input{tables/accu_wbits}

\subsection{Utility and Privacy of Post-Training Quantization}
\label{sec:attack effectiveness}
\smallskip\noindent
{\bf Utility of Post-Training Quantization.}
To gain a comprehensive understanding of 1.58-bit PTQ, we compare the 1.58-bit quantization optimized by AdaRound, BRECQ, and OBC with sign quantization (i.e. 1-bit)~\cite{hubara2018quantized}. The results on CIFAR-100 are summarized in~\Cref{Tab: 1.58-sign} across 10 different random seeds. The accuracy of models quantized to 1.58 bits is significantly higher than that of sign quantization. Compared to the full-precision model, the accuracy drops by only 7.7\%, 2.4\%, and 5.7\% for AdaRound, BRECQ, and OBC, respectively. In contrast, the accuracy of the sign quantized model is similar to random guessing. We find that 1.58-bit quantization provides substantial flexibility and adaptability for PTQs to optimize, making it a practical solution for real-world deployment.

\begin{table}[!h]
\centering
\caption{Test accuracy of extreme low-bit PTQ on CIFAR-100. The full-precision model achieves 70.22\% accuracy. We successfully apply 1.58-bit quantization with the PTQ methods. Sign quantization fails.}
\begin{tabular}{lrrr}
    \hline
        \textbf{Precision} & {\bf AdaRound} & {\bf BRECQ} & {\bf OBC} \\
    \hline
    1.58-bit & 62.35  & 67.61 & 64.34 \\
    1-bit     & 1.00 &  0.99 & 1.04  \\
    \hline
\end{tabular}
\label{Tab: 1.58-sign}  
\end{table}

The 4-bit quantized models achieve accuracy comparable to their full-precision counterparts across all three datasets. In contrast, 2-bit quantization results in a slight accuracy drop. Specifically, according to~\Cref{fig:fig2}, the accuracy degradation of BRECQ 2-bit quantized models is limited to 0.5\%, 0.4\%, and 0.9\% on CIFAR-10, CIFAR-100, and TinyImageNet, respectively. Although the accuracy of 1.58-bit quantization on TinyImageNet is considerably lower than that of the full-precision model (e.g., 33\% with AdaRound 1.58-bit compared to 58\% for the full model), we propose an accuracy-enhancing method and obtain a more usable model in~\Cref{sec:ablation}.

\input{tables/tpr_wbits}

\smallskip\noindent
{\bf Privacy of Post-Training Quantization.}
We observe significant differences in TPR@0.1\%FPR across datasets and attack configurations. In particular, CIFAR-10 shows the lowest TPR values under all methods, indicating its higher resistance to MIA. In contrast, CIFAR-100 and TinyImageNet exhibit significant privacy leakage, with TPR values reaching 37.65\% and 53.04\%, respectively, at 0.1\% FPR. This observation is consistent with prior works~\cite{carlini2022membership,shokri2017membership,zhang2025privacy}, showing that more complex tasks have higher privacy leakage.

Our experiments reveal that extremely low-bit quantization, particularly at 1.58-bit, can reduce TPR@0.1\%FPR across. As shown in~\Cref{fig:fig3}, on CIFAR-100, AdaRound reduces TPR@0.1\%FPR to 0.0031 (Online) and 0.0013 (Offline fixed variance), a reduction of 99.2\% and 99.4\% compared to the full-precision model. On CIFAR-10, OBC achieves values as low as 0.0173 (Online) and 0.0077 (Offline fixed variance), and similar trends are observed for TinyImageNet.

In contrast, 2-bit quantization results in more modest privacy improvements. For instance, AdaRound 2-bit on CIFAR-100 yields a TPR of 0.2196, which is higher than its 1.58-bit counterpart. While 2-bit quantization does reduce TPR@0.1\%FPR, 4-bit quantization often achieves comparable or even higher TPR values, as seen with AdaRound 4-bit on CIFAR-10 (0.0790 vs 0.0776).

The LiRA online attack is more challenging to mitigate than the offline method. For TinyImageNet with 2-bit quantization, the TPR@0.1\% FPR for the Online attack remains at 0.1311 (OBC) and 0.0983 (AdaRound), at the cost of a 9.9\% and 11.1\% decrease in accuracy, respectively. In contrast, offline methods result in near-zero TPR@0.1\%FPR with 2-bit quantization. This discrepancy highlights the importance of using strong attack baselines in privacy research.

\smallskip\noindent
{\bf Efficiency, Utility, and Privacy Trade-off.}
We present the core contribution of our work by analyzing the trade-off between utility and privacy. For easier-to-quantize datasets such as CIFAR-10 and CIFAR-100, 1.58-bit quantization effectively mitigates MIA while maintaining high classification accuracy. On a more challenging dataset, such as TinyImageNet, higher precision (2-bit) is required to preserve comparable accuracy. Although 1.58-bit quantization offers stronger privacy, it causes a noticeable accuracy drop. To address this, we apply 8-bit quantization to the final layer in~\Cref{sec:ablation}, improving performance without compromising privacy.

In general, lower-precision quantization tends to yield better privacy. Across all three datasets, BRECQ 1.58-bit achieves accuracy comparable to AdaRound 2-bit and OBC 2-bit, while consistently obtaining lower TPR@0.1\%FPR. For instance, on TinyImageNet, BRECQ 1.58-bit results in much lower TPR than AdaRound 2-bit, yet with only a 2.01\% drop in accuracy. However, on CIFAR-100, BRECQ 1.58-bit exhibits high variance, rendering the mean TPR insufficient to fully characterize the model’s privacy behavior. Similarly, on CIFAR-10, the TPR under Online attacks also shows noticeable standard deviation. This motivates a deeper analysis of the distribution of accuracy and TPR values.

In summary, our experiments demonstrate that extreme low-bit quantization can significantly reduce TPR@0.1\%FPR with only a modest impact on accuracy. While lower quantization precision typically improves privacy, lower accuracy does not necessarily imply stronger privacy protection. This highlights the importance of carefully selecting both the quantization method and precision level to achieve optimal privacy-utility trade-offs. 
% It is important to emphasize that the conclusions in this section are only preliminary summaries, and further ablation studies in~\Cref{sec:ablation} provide a deeper understanding and practical considerations.

\subsection{Ablation study}
\label{sec:ablation}

\smallskip\noindent
{\bf Decoupled Quantization.}
As discussed in the previous section, due to the significant accuracy degradation caused by 1.58-bit and 2-bit quantization on the TinyImageNet dataset, such low-bit quantization may render the model unusable in scenarios requiring high accuracy.

\begin{figure}[!t]
    \centering
    \begin{minipage}[t]{0.48\linewidth}
        \centering
        \includegraphics[width=\linewidth]{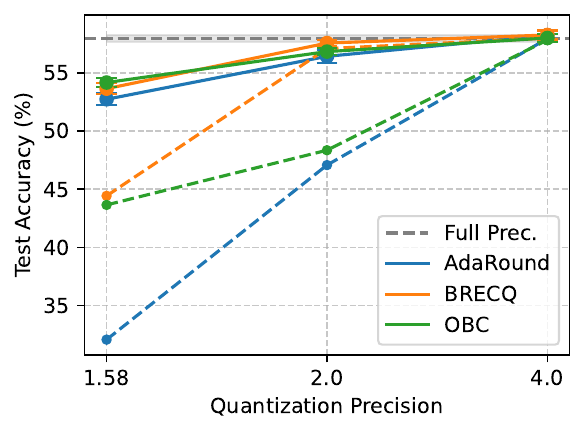}
        \caption*{(a) Test accuracy}
    \end{minipage}
    \hfill
    \begin{minipage}[t]{0.48\linewidth}
        \centering
        \includegraphics[width=\linewidth]{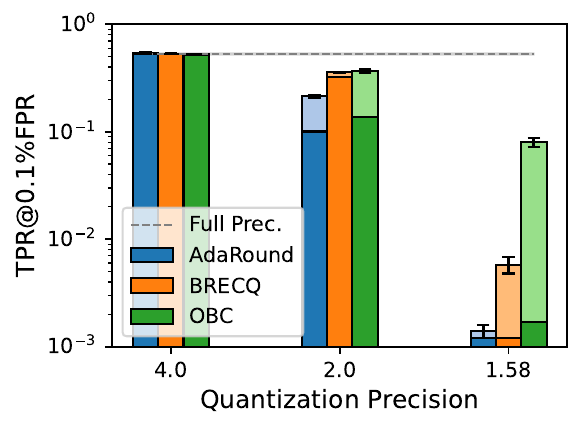}
        \caption*{(b) TPR@0.1\%FPR}
    \end{minipage}    
    \caption{Utility and privacy trade-off of decoupled quantization. Decoupled models control the trade-off by recovering the utility lost at 1.58-bit precision, at the cost of increased privacy leakage. (a) Solid lines use decoupling and dashed lines do not. (b) Each bar consists of the darker bottom representing models without decoupling, and the lighter top with.}
    \label{fig:decoupled_quantizations}
\end{figure}

To address this issue, we employ \emph{decoupled quantization} by quantizing the last layer to a higher precision (i.e. 8 bits) compared to the rest of the model~\cite{li2021brecq}. We study the effect of \emph{decoupled quantization} on the utility-privacy trade-off. \Cref{fig:decoupled_quantizations} (a) shows that after decoupling, the accuracy of the model recovers significantly. For AdaRound (the worst-performing PTQ method), the accuracy drop compared to the full-precision model is reduced to only 1.9\% under 2-bit and 5.6\% under 1.58-bit quantization, compared to 11\% and 26\% before applying the 8-bit last layer, respectively.

The results of the Online attack are presented in~\Cref{fig:decoupled_quantizations} (b), showing that 1.58-bit and 2-bit quantization can still mitigate MIA even after applying 8-bit quantization to the last layer. As test accuracy increases, the privacy leakage measured with TPR@0.1\%FPR also increases. Although the TPR values are higher than in the original low-bit setting, the 1.58-bit quantized models still achieve lower TPR@0.1\%FPR and higher accuracy than the 2-bit models before the decoupled quantization. Our experiments show that higher accuracy does not often correlate with weaker MIA mitigation, as this relationship can vary depending on the quantization method and dataset. Careful selection of quantization strategies can enable a better trade-off between privacy and utility.

\begin{figure}[!t]
    \centering
    \begin{minipage}[t]{0.48\linewidth}
        \centering
        \includegraphics[width=\textwidth]{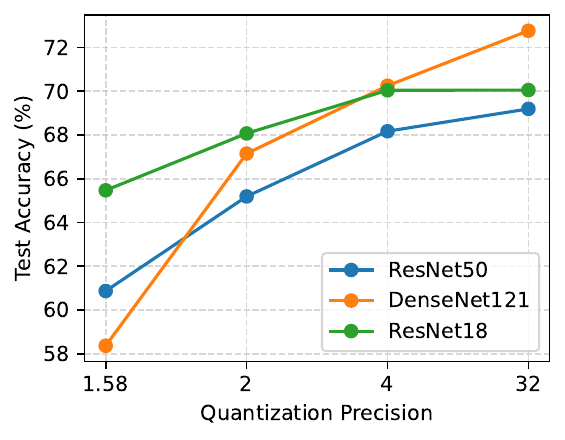}
        \caption*{(a) Test accuracy}
    \end{minipage}
    \hfill
    \begin{minipage}[t]{0.48\linewidth}
        \centering
        \includegraphics[width=\textwidth]{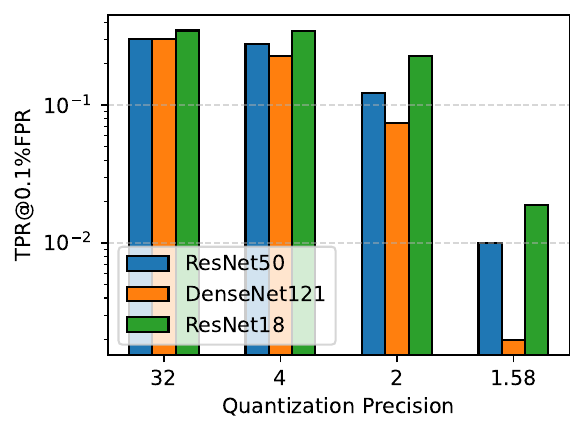}
        \caption*{(b) TPR@0.1\%FPR}
    \end{minipage}
    
    \caption{Utility and privacy across model architectures quantized with OBC on CIFAR-100. We observe a consistent trend for different settings where lower quantization corresponds to a lower accuracy and lower privacy leakage.}
    \label{fig:model_architectures}
\end{figure}

\smallskip\noindent
{\bf Model Architectures.}
We validate our conclusion that the improvement of privacy at the cost of utility through quantization is generalizable and independent of the model architecture. To this end, we use OBC to quantize models with different scales (ResNet18 vs ResNet50), and different family architectures (DenseNet121). We train the target model and 64 shadow models on the same architecture using the CIFAR-100 dataset. 

For quantization, we apply 4-bit, 2-bit, and 1.58-bit OBC quantization to the target model and attack both the full-precision model and the quantized ones. \Cref{fig:model_architectures} shows that all three model architectures exhibit accuracy degradation with a lower precision level. In particular, DenseNet121 has the sharpest decrease among the three architectures. A similar consistent trade-off trend is applied to privacy. The TPR@0.1\%FPR of the 2-bit and 1.58-bit quantized models indicates that, in general, a decrease in accuracy is accompanied by an enhancement in privacy. These findings consistently validate our empirical results across different scales and architectures.

\begin{figure}[!t]
    \centering
    \begin{minipage}[t]{0.48\linewidth}
        \centering
        \includegraphics[width=\linewidth]{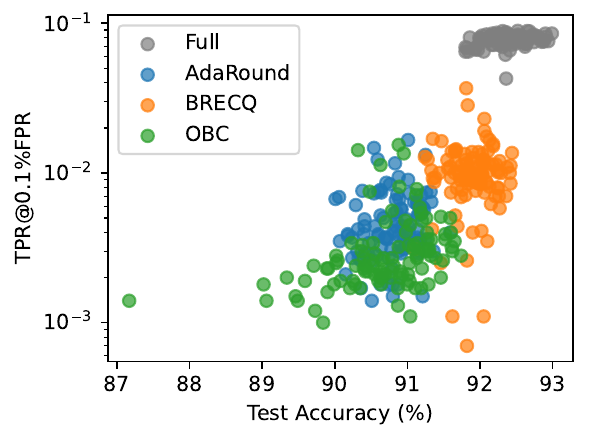}
        \caption*{(a) CIFAR-10}
    \end{minipage}
    \hfill
    \begin{minipage}[t]{0.48\linewidth}
        \centering
        \includegraphics[width=\linewidth]{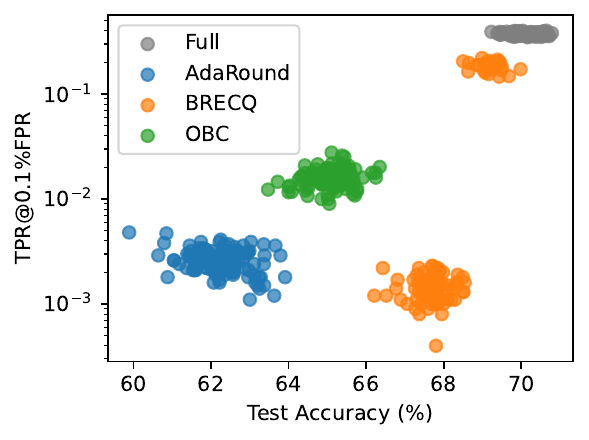}
        \caption*{(b) CIFAR-100}
    \end{minipage}
    \caption{Utility and privacy analysis for 1.58-bit quantized models across 100 runs. For CIFAR-100, AdaRound and OBC exhibit a concentrated distribution. In contrast, BRECQ shows two distinct clusters: some models have significantly higher privacy leakage, together with higher accuracy. Among all methods, BRECQ consistently achieves the highest accuracy and the lowest privacy leakage.}
    \label{fig:scatter}
\end{figure}

\smallskip\noindent
{\bf Variance in 1.58-bits Models.}
A significant privacy discrepancy, measured with TPR@0.1\%, is observed for BRECQ quantized models at 1.58 bits. This has been previously observed by Carlini et al., 2022~\cite{carlini2022membership}, reporting a large variance in the privacy of non-quantized models with similar utility. We observe the same phenomenon for low-bit quantized models.
To analyze the robustness and privacy characteristics of 1.58-bit quantized models, we visualize the trade-off between accuracy and privacy risk on both CIFAR-100 and CIFAR-10 datasets in~\Cref{fig:scatter}.

On CIFAR-100, BRECQ consistently achieves higher accuracy compared to AdaRound and OBC. However, its TPR@0.1\%FPR exhibits significant variance, indicating unstable privacy leakage. Importantly, we observe that BRECQ also achieves the lowest TPR values among all methods, suggesting that it is capable of delivering the strongest privacy protection under certain conditions. In contrast, on CIFAR-10, all methods, including BRECQ, exhibit less variance in privacy leakage. However, BRECQ still shows a wider spread in TPR@0.1\%FPR compared to AdaRound and OBC. 

% These results suggest that the accuracy–privacy trade-off is more sensitive to quantization strategies on datasets with higher class complexity (like CIFAR-100). In particular, BRECQ’s optimization process may amplify small differences in model behavior that substantially affect privacy risks. 
% \hltext{Although a large variance is observed, the results of our experiment still offer valuable insights. In most cases, the TPR@0.1\%FPR is concentrated around the mean or exceeds it (e.g., BRECQ on CIFAR-100).} 

%%%%%%%%%%%%%%%%%%%%%%%%%%%%%%%%%%%%%%%%%%%%%%%%%%%%%%%%%%%%%%%%%%%%%%%%

\section{Related Work}
\label{sec:related}

\smallskip\noindent
Quantization, as a model compression technique, can smooth model outputs and reduce the influence of outliers. At the same time, the precision degradation caused by quantization inherently introduces noise into the model parameters, which is considered to provide privacy protection for the model~\cite{towards2022model}. Early research on model compression and privacy, such as the method proposed by Wang et al., 2019~\cite{wang2019private} to enhance differential privacy through knowledge distillation, achieves meaningful differential privacy with high utility; Mia et al., 2023~\cite{mia2023compression} conduct experiments on various model compression techniques to explore their impact on privacy. However, these works were limited to non-practical quantization techniques that do not achieve high-accuracy low-bit PTQ.

In recent years, several works have added quantization during model training to enhance privacy protection. For example,
incorporating quantization as part of their methods in federated learning~\cite{zhu2021distributed,he2020cossgd,lang2023joint,ZHANG2021106775}, and combining quantization with differential privacy~\cite{kang2024effect,amiri2021compressive}. 
These studies suggest that quantization can offer a certain degree of privacy protection. % however, they either combine quantization with additional techniques such as differential privacy, or apply it only during model training, limiting their applicability to specific scenarios.

Deng et al., 2025~\cite{deng2025private} utilize information theory to analyze the privacy offered by quantized models, while Amiri et al., 2021~\cite{amiri2021compressive} rank the privacy levels of various quantization procedures. These works theoretically demonstrate the privacy-preserving effect of quantization. The analyses are primarily applicable to simple quantization and smaller models. For more complex PTQ and deep neural networks, theoretical analysis remains a challenge.

As a fundamental and long-studied attack method, MIA has been widely applied in model privacy analysis. Yang et al., 2024~\cite{yang2024securing} propose a fine-tuned quantization method to defend against MIA, where they quantize the final layer of the model while simulating MIA and attempting to reduce its accuracy. Their approach successfully reduces MIA accuracy while maintaining test accuracy.  However, it does not explicitly isolate or quantify the contribution of quantization itself to the model’s robustness against such attacks. Famili et al., 2023~\cite{famili2023deep} improve the privacy leakage to MIA by using quantization-aware training, demonstrating notable improvements over full-precision models. These findings indicate that QAT can effectively mitigate MIA; the impact of PTQ remains largely unexamined. Due to the practical advantages and widespread use of PTQ in real-world deployments, we carry out a comprehensive investigation into its effect on MIA vulnerability.

%%%%%%%%%%%%%%%%%%%%%%%%%%%%%%%%%%%%%%%%%%%%%%%%%%%%%%%%%%%%%%%%%%%%%%%%

\section{Discussion}
\smallskip\noindent
In this paper, we systematically evaluate privacy leakage across different post-training quantization (PTQ) methods using the state-of-the-art LiRA membership inference attack. We conduct our study in a controlled experimental setting that enables precise manipulation of training conditions and exhaustive evaluation. This helps to better understand findings in prior large-scale studies with limited configurations~\cite{mia2023compression}. 

Our results extends prior findings on standard quantization methods~\cite{yang2024securing,famili2023deep}, while adding the analysis to more aggressive quantization schemes. We observe that 4-bit quantization preserves both model utility and privacy leakage at levels comparable to full-precision models. However, further reducing precision to 2-bit and 1.58-bit reveals a privacy-utility trade-off: \emph{model accuracy degrades while privacy protection increases}, as measured by reduced membership inference attack success rates. These findings demonstrate that quantization precision can serve as a tunable parameter for balancing privacy and utility requirements, with practical implications for deploying efficient models under privacy constraints.

Several future directions can be built upon on our findings. For example, it is possible to straightforwardly expand the analysis to other domains, such as text or graphs. 
Alternatively, one can include the activation quantization in the analysis, and/or combine quantization with other privacy defenses~\cite{zhu2021distributed}.
Lastly, following the large-scale direction, which requires extensive computation, one can study the privacy of models trained with and without quantization under the scaling law \cite{kaplan2020scaling}. 

\smallskip\noindent
{\bf Impact.}
Our results are at the intersection of efficient and private machine learning. We empirically quantify the privacy leakage of state-of-the-art post-trained quantized methods, offering practical insights in choosing the appropriate trade-off between utility, privacy, and quantization levels. 

\smallskip\noindent
{\bf Limitations.}
Our results are limited to the standard vision benchmark commonly used in the privacy community. The conclusions are based on empirical evidence observed via controlled experiments using post-training quantization, instead of analytical results. Furthermore, balancing privacy and utility at deployment is application-specific, and it can require extensive parameter tuning.

\bibliographystyle{IEEEtran}
\bibliography{mybibfile}
% -----------------------------------------------------------------------
% -----------------------------------------------------------------------
% -----------------------------------------------------------------------

\end{document}

%% file: tables/accu_wbits.tex
\begin{figure*}[!t]
    \centering
    \begin{minipage}{0.30\textwidth}
        \centering
        \includegraphics[width=\linewidth]{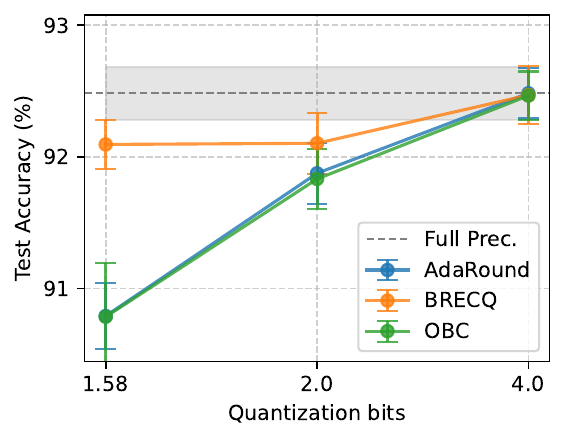}
        \caption*{(a) CIFAR-10}
        \label{fig:sub1}
    \end{minipage}
    \hfill
    \begin{minipage}{0.30\textwidth}
        \centering
        \includegraphics[width=\linewidth]{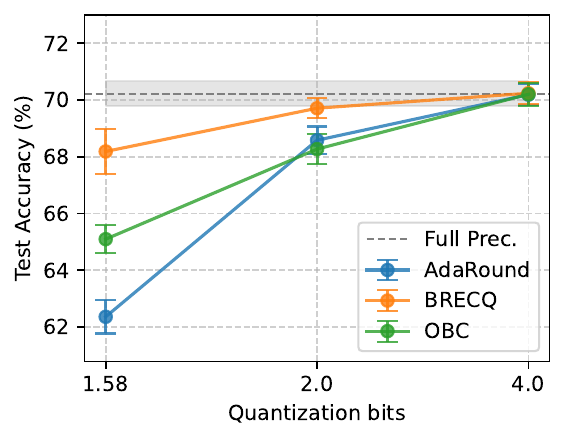}
        \caption*{(b) CIFAR-100}
        \label{fig:sub2}
    \end{minipage}
    \hfill
    \begin{minipage}{0.30\textwidth}
        \centering
        \includegraphics[width=\linewidth]{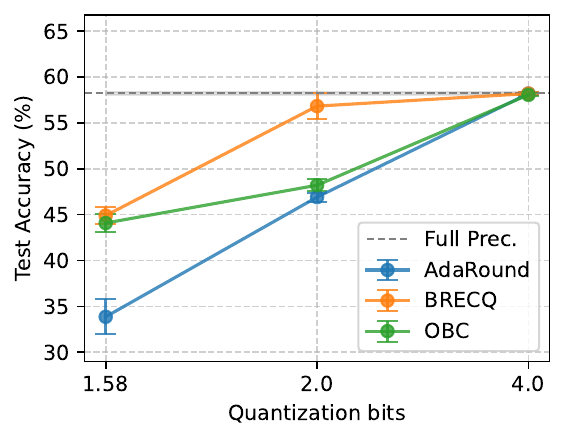}
        \caption*{(c) TinyImageNet}
        \label{fig:sub3}
    \end{minipage}

    \caption{Test accuracy of the models quantized with different PTQs at various precision levels. As expected, we observe that lower precision corresponds to lower accuracy. In particular, at the 1.58-bit level, the utility drop becomes significant. Results are averaged over 5 random seeds.}
    \label{fig:fig2}
\end{figure*}

%% file: tables/tpr_wbits.tex
\begin{figure*}[!th]
    \centering
    \begin{minipage}{0.30\textwidth}
        \centering
        \includegraphics[width=\linewidth]{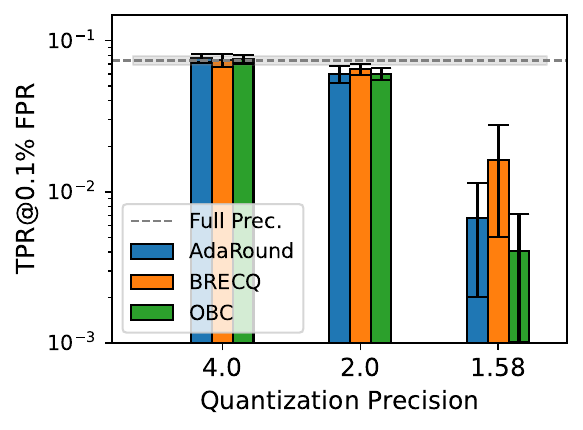}
        \caption*{(a) CIFAR-10, Online}
        \label{fig:cifar10-tpr-online}
    \end{minipage}
    \hfill
    \begin{minipage}{0.30\textwidth}
        \centering
        \includegraphics[width=\linewidth]{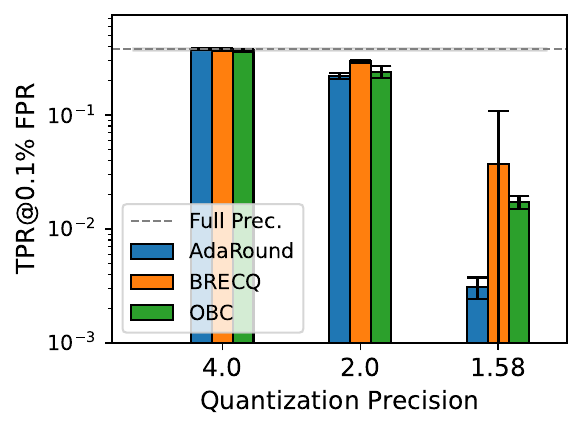}
        \caption*{(e) CIFAR-100, Online}
        \label{fig:cifar100-tpr-online}
    \end{minipage}
    \hfill
    \begin{minipage}{0.30\textwidth}
        \centering
        \includegraphics[width=\linewidth]{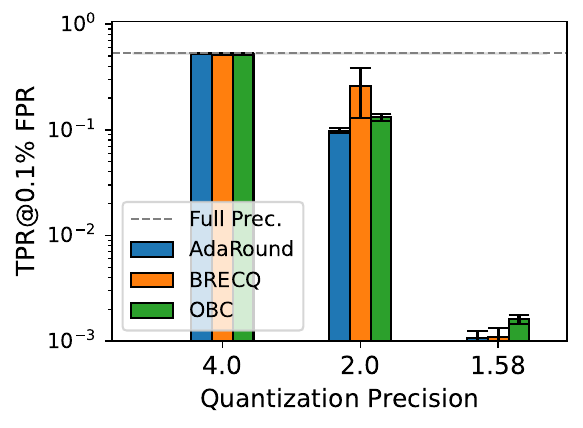}
        \caption*{(c) TinyImageNet, Online}
        \label{fig:tiny-tpr-online}
    \end{minipage}

    \begin{minipage}{0.30\textwidth}
        \centering
        \includegraphics[width=\linewidth]{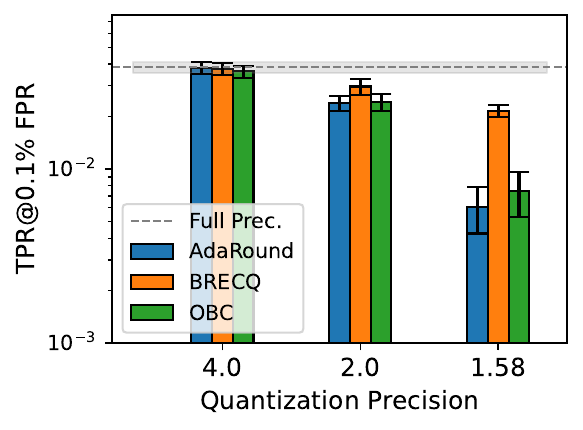}
        \caption*{(d) CIFAR-10, Offline}
        \label{fig:cifar10-tpr-offline}
    \end{minipage}
    \hfill
    \begin{minipage}{0.30\textwidth}
        \centering
        \includegraphics[width=\linewidth]{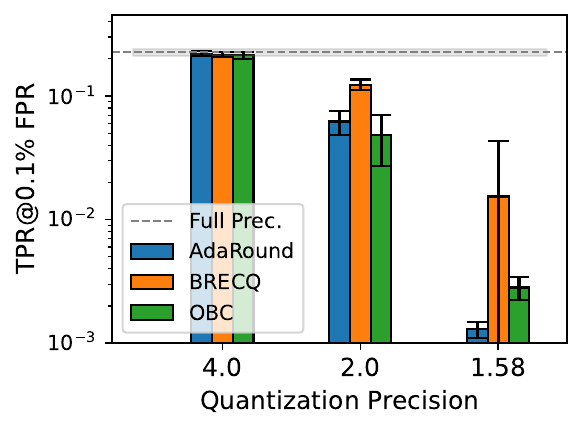}
        \caption*{(b) CIFAR-100, Offline}
        \label{fig:cifar100-tpr-offline}
    \end{minipage}
    \hfill
    \begin{minipage}{0.30\textwidth}
        \centering
        \includegraphics[width=\linewidth]{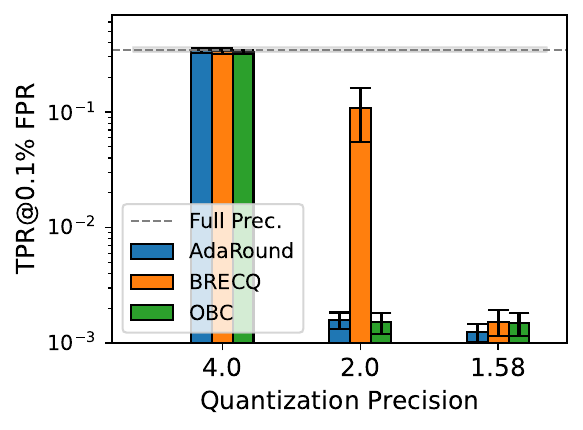}
        \caption*{(f) TinyImageNet, Offline}
        \label{fig:tiny-tpr-offline}
    \end{minipage}

    \caption{Privacy leakage of the models quantized with different PTQs at various precision levels, under the LiRA online and LiRA offline attack. While 4-bit quantization has similar privacy as the full precision, 2-bit and 1.58-bit can consistently mitigate the privacy leakage. In particular, the 1.58-bit quantization significantly reduces the MIA success.}
    \label{fig:fig3}
\end{figure*}

%% file: mybibfile.bib
@incollection{gholami2022survey,
  title={A survey of quantization methods for efficient neural network inference},
  author={Gholami, Amir and Kim, Sehoon and Dong, Zhen and Yao, Zhewei and Mahoney, Michael W and Keutzer, Kurt},
  booktitle={Low-power Computer Vision},
  pages={291--326},
  year={2022},
}

@article{nagel2021white,
  title={A white paper on neural network quantization},
  author={Nagel, Markus and Fournarakis, Marios and Amjad, Rana Ali and Bondarenko, Yelysei and Van Baalen, Mart and Blankevoort, Tijmen},
  journal={arXiv preprint arXiv:2106.08295},
  year={2021}
}

@article{le2023efficient,
  title={Efficient neural networks for tiny machine learning: A comprehensive review},
  author={L{\^e}, Minh Tri and Wolinski, Pierre and Arbel, Julyan},
  journal={arXiv preprint arXiv:2311.11883},
  year={2023}
}

@inproceedings{nagel2020up,
  title={Up or down? adaptive rounding for post-training quantization},
  author={Nagel, Markus and Amjad, Rana Ali and Van Baalen, Mart and Louizos, Christos and Blankevoort, Tijmen},
  booktitle={Proceedings of International Conference on Machine Learning},
  pages={7197--7206},
  year={2020},
}

@inproceedings{
li2021brecq,
title={{\{}BRECQ{\}}: Pushing the Limit of Post-Training Quantization by Block Reconstruction},
author={Yuhang Li and Ruihao Gong and Xu Tan and Yang Yang and Peng Hu and Qi Zhang and Fengwei Yu and Wei Wang and Shi Gu},
booktitle={Proceedings of International Conference on Learning Representations},
year={2021},
}

@article{frantar2022optimal,
  title={Optimal brain compression: A framework for accurate post-training quantization and pruning},
  author={Frantar, Elias and Alistarh, Dan},
  journal={Advances in Neural Information Processing Systems},
  volume={35},
  pages={4475--4488},
  year={2022}
}

@article{ma2024era,
  title={The era of 1-bit llms: All large language models are in 1.58 bits},
  author={Ma, Shuming and Wang, Hongyu and Ma, Lingxiao and Wang, Lei and Wang, Wenhui and Huang, Shaohan and Dong, Lifeng and Wang, Ruiping and Xue, Jilong and Wei, Furu},
  journal={arXiv preprint arXiv:2402.17764},
  year={2024},
}

@inproceedings{shokri2017membership,
  title={Membership inference attacks against machine learning models},
  author={Shokri, Reza and Stronati, Marco and Song, Congzheng and Shmatikov, Vitaly},
  booktitle={Proceedings of the 2017 IEEE Symposium on Security and Privacy},
  pages={3--18},
  year={2017},
}

@InProceedings{HP23,
  author    = {H. Hu and J. Pang},
  title     = {Loss and likelihood based membership inference of diffusion models},
  booktitle = {Proceedings of the 26th International Conference on Information Security},
  year      = {2023},
  pages     = {121-141},
}

@inproceedings{carlini2022membership,
  title={Membership inference attacks from first principles},
  author={Carlini, Nicholas and Chien, Steve and Nasr, Milad and Song, Shuang and Terzis, Andreas and Tramer, Florian},
  booktitle={Proceedings of the 2022 IEEE Symposium on Security and Privacy},
  pages={1897--1914},
  year={2022},
}

@inproceedings{towards2022model,
  title={Towards model quantization on the resilience against membership inference attacks},
  author={Kowalski, Charles and Famili, Azadeh and Lao, Yingjie},
  booktitle={IEEE International Conference on Image Processing},
  pages={3646--3650},
  year={2022},
}

@article{mia2023compression,
  title={Membership Inference Attacks Against Compression Models},
  author={Chen, J. and Sun, K. and Zhao, L.},
  journal={Computing Journal},
  year={2023},
}

@article{deng2025private,
  title={Private inference in quantized models},
  author={Deng, Zirui and Ramkumar, Vinayak and Bitar, Rawad and Raviv, Netanel},
  journal={IEEE Transactions on Information Theory},
  year={2025},
  publisher={IEEE}
}

@article{dnn2023quantization,
  title={Deep Neural Network Quantization Framework for Effective Defense Against Membership Inference Attacks},
  author={Chen, B. and Zhang, H. and Liu, X.},
  journal={Sensors},
  volume={23},
  number={18},
  pages={7722},
  year={2023},
}

@article{lang2023joint,
  title={Joint privacy enhancement and quantization in federated learning},
  author={Lang, Natalie and Sofer, Elad and Shaked, Tomer and Shlezinger, Nir},
  journal={IEEE Transactions on Signal Processing},
  volume={71},
  pages={295--310},
  year={2023},
}

@inproceedings{kang2024effect,
  title={The effect of quantization in federated learning: a r{\'e}nyi differential privacy perspective},
  author={Kang, Tianqu and Liu, Lumin and He, Hengtao and Zhang, Jun and Song, SH and Letaief, Khaled B},
  booktitle={Proceedings of the 2024 IEEE International Mediterranean Conference on Communications and Networking },
  pages={233--238},
  year={2024},
}

@article{zhu2021distributed,
  title={Distributed additive encryption and quantization for privacy preserving federated deep learning},
  author={Zhu, Hangyu and Wang, Rui and Jin, Yaochu and Liang, Kaitai and Ning, Jianting},
  journal={Neurocomputing},
  volume={463},
  pages={309--327},
  year={2021},
}

@article{yeom2018privacy,
  author    = {Samuel Yeom and Matthew Fredrikson and Somesh Jha},
  title     = {Privacy Risk in Machine Learning: Analyzing the Connection to Overfitting},
  journal   = {Proceedings of the IEEE Computer Security Foundations Symposium},
  year      = {2018},
  pages     = {268-282},
}

@article{hubara2018quantized,
  title={Quantized neural networks: Training neural networks with low precision weights and activations},
  author={Hubara, Itay and Courbariaux, Matthieu and Soudry, Daniel and El-Yaniv, Ran and Bengio, Yoshua},
  journal={Journal of Machine Learning Research},
  volume={18},
  number={187},
  pages={1--30},
  year={2018}
}

@article{jin2023membership,
  title={Membership inference attacks against compression models},
  author={Jin, Yong and Lou, Weidong and Gao, Yanghua},
  journal={Computing},
  volume={105},
  number={11},
  pages={2419--2442},
  year={2023},
}

@inproceedings{amiri2021compressive,
  title={Compressive differentially private federated learning through universal vector quantization},
  author={Amiri, Saba and Belloum, Adam and Klous, Sander and Gommans, Leon},
  booktitle={Proceedings of the AAAI Workshop on Privacy-Preserving Artificial Intelligence},
  pages={2--9},
  year={2021}
}

@inproceedings{yang2024securing,
  title={Securing Deep Neural Networks on Edge from Membership Inference Attacks Using Trusted Execution Environments},
  author={Yang, Cheng-Yun and Ramshankar, Gowri and Eliopoulos, Nicholas and Jajal, Purvish and Nambiar, Sudarshan and Miller, Evan and Zhang, Xun and Tian, Dave and Chen, Shuo-Han and Perng, Chiy-Ferng and others},
  booktitle={Proceedings of the 29th ACM/IEEE International Symposium on Low Power Electronics and Design},
  pages={1--6},
  year={2024}
}

@article{famili2023deep,
  title={Deep neural network quantization framework for effective defense against membership inference attacks},
  author={Famili, Azadeh and Lao, Yingjie},
  journal={Sensors},
  volume={23},
  number={18},
  pages={7722},
  year={2023},
}

@article{ZHANG2021106775,
title = {A survey on federated learning},
journal = {Knowledge-Based Systems},
volume = {216},
pages = {106775},
year = {2021},
author = {Chen Zhang and Yu Xie and Hang Bai and Bin Yu and Weihong Li and Yuan Gao},
}

@article{he2020cossgd,
  title={CosSGD: Nonlinear quantization for communication-efficient federated learning},
  author={He, Yang and Zenk, Maximilian and Fritz, Mario},
  journal={arXiv preprint arXiv:2012.08241},
  year={2020}
}

@techreport{krizhevsky2009learning,
  title={Learning multiple layers of features from tiny images},
  author={Krizhevsky, Alex},
  year={2009},
  institution={University of Toronto}
}

@article{le2015tiny,
  title={Tiny imagenet visual recognition challenge},
  author={Le, Yann and Yang, Xuan},
  journal={CS231N Stanford},
  volume={7},
  number={7},
  pages={3},
  year={2015}
}

@inproceedings{he2016deep,
  title={Deep residual learning for image recognition},
  author={He, Kaiming and Zhang, Xiangyu and Ren, Shaoqing and Sun, Jian},
  booktitle={Proceedings of the IEEE Conference on Computer Vision and Pattern Recognition},
  pages={770--778},
  year={2016}
}

@article{russakovsky2015imagenet,
  title={ImageNet Large Scale Visual Recognition Challenge},
  author={Russakovsky, Olga and Deng, Jia and Su, Hao and Krause, Jonathan and Satheesh, Sanjeev and Ma, Sean and Huang, Zhiheng and Karpathy, Andrej and Khosla, Aditya and Bernstein, Michael and Berg, Alexander C and Fei-Fei, Li},
  journal={International Journal of Computer Vision},
  volume={115},
  number={3},
  pages={211--252},
  year={2015},
}

@inproceedings{tang2022mitigating,
  title={Mitigating membership inference attacks by self-distillation through a novel ensemble architecture},
  author={Tang, Xinyu and Mahloujifar, Saeed and Song, Liwei and Shejwalkar, Virat and Nasr, Milad and Houmansadr, Amir and Mittal, Prateek},
  booktitle={Proceedings of the 31st USENIX Security Symposium},
  pages={1433--1450},
  year={2022}
}

@inproceedings{abadi2016deep,
  title={Deep learning with differential privacy},
  author={Abadi, Martin and Chu, Andy and Goodfellow, Ian and McMahan, H Brendan and Mironov, Ilya and Talwar, Kunal and Zhang, Li},
  booktitle={Proceedings of the 2016 ACM SIGSAC Conference on Computer and Communications Security},
  pages={308--318},
  year={2016}
}

@inproceedings{zhang2017understanding,
  title={Understanding deep learning requires rethinking generalization},
  author={Zhang, Chiyuan and Bengio, Samy and Hardt, Moritz and Recht, Benjamin and Vinyals, Oriol},
  booktitle={Proceedings of the International Conference on Learning Representations},
  year={2017}
}

@inproceedings{wang2019private,
  title={Private model compression via knowledge distillation},
  author={Wang, Ji and Bao, Weidong and Sun, Lichao and Zhu, Xiaomin and Cao, Bokai and Yu, Philip S},
  booktitle={Proceedings of the AAAI Conference on Artificial Intelligence},
  pages={1190--1197},
  year={2019}
}

@article{hooker2019compressed,
  title={What do compressed deep neural networks forget?},
  author={Hooker, Sara and Courville, Aaron and Clark, Gregory and Dauphin, Yann and Frome, Andrea},
  journal={arXiv preprint arXiv:1911.05248},
  year={2019}
}

@article{hooker2020characterising,
  title={Characterising bias in compressed models},
  author={Hooker, Sara and Moorosi, Nyalleng and Clark, Gregory and Bengio, Samy and Denton, Emily},
  journal={arXiv preprint arXiv:2010.03058},
  year={2020}
}

@article{zhang2025privacy,
  title={Spurious Privacy Leakage in Neural Networks},
  author={Zhang, Chenxiang and Pang, Jun and Mauw, Sjouke},
  journal={Transactions on Machine Learning Research},
  year={2025}
}

@article{kaplan2020scaling,
  title={Scaling laws for neural language models},
  author={Kaplan, Jared and McCandlish, Sam and Henighan, Tom and Brown, Tom B and Chess, Benjamin and Child, Rewon and Gray, Scott and Radford, Alec and Wu, Jeffrey and Amodei, Dario},
  journal={arXiv preprint arXiv:2001.08361},
  year={2020}
}
